\documentclass[preprint,journal]{vgtc}       %

\ifpdf%
  \pdfoutput=1\relax                   %
  \usepackage{graphicx}                %
  \DeclareGraphicsExtensions{.pdf,.png,.jpg,.jpeg} %
\else%
  \usepackage{graphicx}                %
  \DeclareGraphicsExtensions{.eps}     %
\fi%

\graphicspath{{figures/}{pictures/}{images/}{./}} %

\usepackage{microtype}                 %
\PassOptionsToPackage{warn}{textcomp}  %
\usepackage{textcomp}                  %
\usepackage{mathptmx}                  %
\usepackage{times}                     %
\usepackage{cite}                      %
\usepackage{tabu}                      %
\usepackage{booktabs}                  %
\usepackage{hyperref}

\onlineid{0}

\vgtccategory{Research}
\vgtcpapertype{please specify}
\usepackage{xspace}
\newcommand{\model}{PromptIDE\xspace}

\title{Interactive and Visual Prompt Engineering \\ for Ad-hoc 
Task Adaptation with Large Language Models }

\author{Hendrik Strobelt, Albert Webson, Victor Sanh, Benjamin Hoover, \\Johanna Beyer, Hanspeter Pfister, and Alexander M. Rush}
\authorfooter{
 H. Strobelt and B. Hoover are with IBM Research.
 A. Webson is with Brown University. 
 V. Sanh and A. Rush are with Huggingface. 
 J. Beyer and H. Pfister are with Harvard SEAS. 
 A. Rush is with Cornell Tech.
}

\shortauthortitle{Biv \MakeLowercase{\textit{et al.}}: Global Illumination for Fun and Profit}

\abstract{
State-of-the-art neural language models can now be used to solve ad-hoc language tasks through zero-shot \textit{prompting} without the need for supervised training. This approach has gained popularity in recent years, and researchers have demonstrated prompts that achieve strong accuracy on specific NLP tasks. However, finding a prompt for new tasks requires experimentation. Different prompt templates with different wording choices lead to significant accuracy differences. \model allows users to experiment with prompt variations, visualize prompt performance, and iteratively optimize prompts. We developed a workflow that allows users to first focus on model feedback using small data before moving on to a large data regime that allows empirical grounding of promising prompts using quantitative measures of the task. The tool then allows easy deployment of the newly created ad-hoc models. We demonstrate the utility of \model (demo: \url{http://prompt.vizhub.ai}) and our workflow using several real-world use cases. 
} %


\teaser{
  \centering
  \includegraphics[width=.9\linewidth]{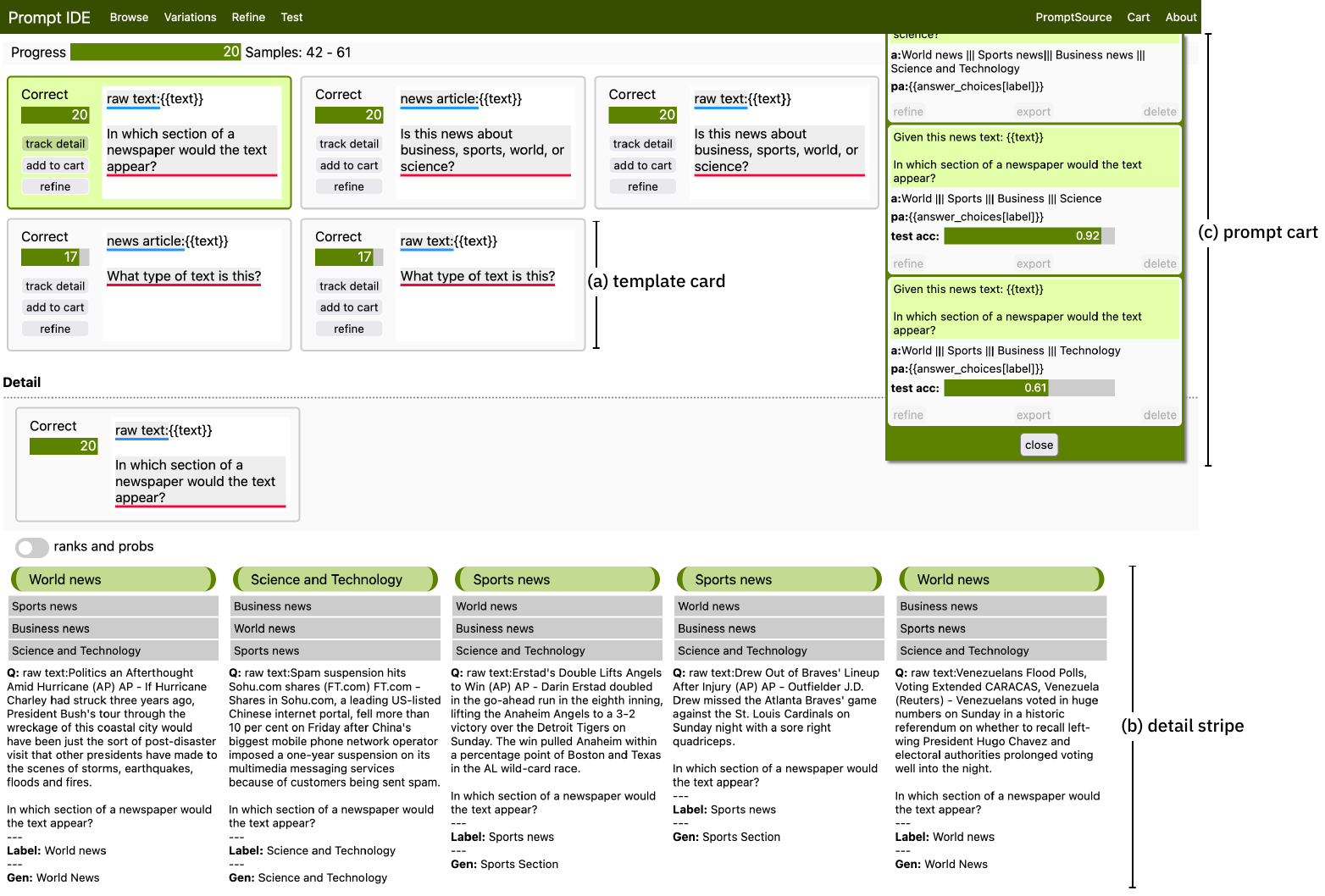}
  \caption{Example of \model's interface to explore variations of different prompts. Each variation is tested against up to twenty data examples and represented as a template card (a). For each variation, rich detail can be tracked by using the detail stripes (b). If performance and qualitative detail are convincing, a user can collect the prompt in the prompt cart (c).  }
  \label{fig:teaser}
}

\vgtcinsertpkg

\begin{document}

\firstsection{Introduction}

\maketitle
Machine learning models for natural language processing (NLP) have shown impressive results on benchmark tasks; however, translating this success from model architects into specific applications for model users remains a challenge. One challenge is that benchmarks typically assume a supervised train-test workflow. The underlying task is carefully designed top-down with annotated training data. However, a significant portion of use-cases of NLP does not easily fit this workflow. For example, consider a journalist, covering legal proceedings, who is interested in finding all cited instances of a precedent~\cite{popescu2017natural}, or a financial analyst looking through past company financial statements to find cases of debt obligations~\cite{el2020finsim}. Annotating enough data, training a model, and then applying it to their task requires time and expertise that may not be available for a user in this setting.

In recent years, an alternative bottom-up approach, known as \textit{prompting}, has become popular for developing ad-hoc end-user tasks in NLP~\cite{radford2019language,brown2020gpt3,schick-schutze-2021-just}.
To solve an ad-hoc task, the user provides, in natural language, a prompt template that describes the task along with target answer choices. For example, simply providing a prompt template such ``Is the case \{case\} referenced in this text? \{text\}'' could alone provide a classification model for an ad-hoc task with no explicit train and test data needed. This approach is possible through advances in training large general-purpose models for language. 

The promise of prompting is that it allows domain experts to solve new tasks with only natural language inputs. However, while there are prompts that can achieve high accuracy on specific tasks, there is a large amount of variance in the choice of the prompt template itself.  Recent papers have described how  task accuracy is dependent on specifics of prompt choices~\cite{perez:neuraips2021,zhao:arxiv2021,albert:arxiv2021}. This leads to a brute-force procedure under which dozens of prompts are written, evaluated, and compared to find the best fit for a task. In this sense, prompting transfers similar burdens of curating expert labels to prompt construction. 

This work explores how interactive visualization can support prompt construction for domain experts. Unlike aspects of model training, such as hyperparameter tuning, prompting is not constrained to brute-force exploration. Because prompt templates are written in natural language, users can craft and customize them for their tasks of interest and refine their answer choices based on the dataset. Outputs are legible to domain experts who can observe the process to stop it early or adjust it based on failures. They can rewrite prompts based on system observation to find the best expressions for their task. The tool is agnostic to the underlying model or datasets used but aims to support the expert in their goals.

This work makes the following novel contributions. 
(1) \model automates the creation and evaluation of many prompt templates simultaneously and supports different underlying models, tasks, and datasets.
(2) \model encourages a principled and repeatable workflow for prompt engineering. Users are guided through the process, with opportunities for iterations at each step. 
(3) We demonstrate the utility of \model and our workflow for several real-world use cases for ad-hoc NLP models. We end the paper by exploring future challenges and avenues for follow-up work.

\section{Related Work}
\label{sec:relwork}
\subsection{Prompting as an Interface}

The use of large language models as a replacement for supervised learning was popularized by the GPT
series of language models~\cite{radford2019language}. Prompting both in the zero-short and few-shot settings has been explored widely in NLP tasks~\cite{schick-schutze-2021-just,le-scao-rush-2021-many,gao-etal-2021-making,wei2022chain}. We consider only the case of human-readable prompts to large models, which contrasts with methods such as prompt-tuning~\cite{DBLP:conf/acl/LiL20, DBLP:conf/emnlp/LesterAC21}, which learns a continuous prompt embedding, and auto-prompting~\cite{DBLP:conf/emnlp/ShinRLWS20}, which attempts to search for prompts from scratch, both of which require a training step.  For more examples see a recent survey by Liu et al.~\cite{liu:corr2021}. Current prompt-based models are primarily based on Transformers~\cite{vaswani2017attention}, which has become the de-facto model architecture for NLP models; however, nothing about our visual analysis is specific to the model used or task considered, only the prompting interface.

Prompting as a means to interact with generative models is prevalent in online demos that often deploy Transformers as assistive agents. For instance, OpenAI released a (closed-source) API with a simple demo to interact with their proprietary GPT-3 model using text~\cite{openai2022playground}.\footnote{Many demos built around this API can be found at \url{https://gptcrush.com/}}
Another notable example is Github CoPilot~\cite{githubCopilot}, which uses OpenAI's Codex~\cite{chen2021codex}, a GPT-like language model trained on code, to allow everyday programmers to turn natural language prompts into code through an IDE like VSCode. With this model, a handful of well-crafted natural language instructions can code entirely functional browser-based video games~\cite{mayne2022building}.
The flexibility and effectiveness of prompt-based approaches encourage the use of prompting as the preferred way to interact with powerful NLP models.

The flexibility of prompts as an interface also comes with a cost, as downstream performance is closely tied to prompt wording. Writing good prompts to extract the desired information from a model is usually left to trial and error by the user, with few exceptions.
One tool that seeks to solve this problem is Prompts.ai~\cite{zhidkov2020prompts}, a visual tool that uses the GPT-3 API to explore how a user-specified prompt template affects the behavior of GPT-3 on individual examples, in conversations, and with different generation parameters. However, its input space is limiting, allowing only one prompt template with limiting syntax to be tested at a time.
PromptSource~\cite{bach2022promptsource} increases the power of the templating language used to write general prompts.
It also provides a platform for the community to create, evaluate, and explore new prompts. \model extends this work, providing a principled workflow to automate the time-intensive process of creating and evaluating many prompt templates on different models, datasets, and tasks.

\subsection{Visualization for NLP}
Visualization tools to interact with language models have grown increasingly popular alongside the rise of the models themselves. These tools can serve several functions: (1) to expose the internals of a particular architecture (e.g., Transformers, RNNs, LSTMs) to understand how it encodes knowledge of the language~\cite{hoover2020exbert,vig2019bertviz,alammar2020explaining,wang2021dodrio,strobelt2017lstmvis,derose2020attention,li2021t3,ming2017understanding,jaunet2021visqa}; (2) to explore and compare the behavior of a model’s internal distributions during text generation~\cite{gehrmann2019gltr,strobelt2021lmdiff}; and (3) to understand how model behavior differs under controlled input or parameter changes~\cite{tenney2020language,pearce2021language,liu2018nlize}. \model is agnostic to the underlying language model and thus serves the functions of (2) and (3).

Other visualization tools that treat NLP models as black boxes focus on visualizations of output distributions and model performance with single custom input or static sets of inputs. 
For example, GLTR~\cite{gehrmann2019gltr} focuses on visualizing output probability distributions to support humans in detecting whether a provided text was generated by a model. LMdiff~\cite{strobelt2021lmdiff} extends these visualizations to support comparisons between different language models on user-provided inputs and static corpora.
Neither of these provides an exploration of the input space.
The Language Interpretability Tool (LIT)~\cite{tenney2020language} is a comprehensive toolkit that enables rapid exploration and error analysis for a model on a larger input dataset. However, LIT's comprehensive analyses are not conducive for rapid, iterative improvements of a prompt on general NLP tasks. NLIZE~\cite{liu2018nlize} serves as a debugging tool that evaluates how a language model outputs changes as a result of perturbations to its hidden state rather than its input. Unlike these existing tools, \model enables exploration and evaluation of the infinitely large space of possible input prompts. 

\section{Model: Prompting for NLP}
\label{sec:model}

\begin{figure}[tb]
    \centering
    \includegraphics[width=\linewidth]{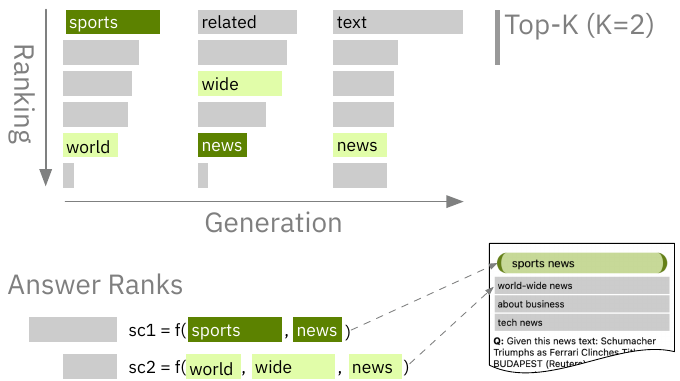}
    \caption{A language model can \textbf{generate} text by \textbf{ranking} all possible events (tokens) at each time step based on their probability. Often, only \textbf{top-k} tokens are considered as generation output. In \model, only the \textbf{answer ranks} of defined answer choices are compared against each other.}
    \label{fig:formulas_explained}
\end{figure}

Prompting is a paradigm for solving ad-hoc NLP tasks. We particularly focus on zero-shot prompting that assumes we do not have access to any training examples. Prompting is typically used in conjunction with large pre-trained language models~\cite{radford2019language,brown2020gpt3}. These models are powerful, but their size makes them difficult to train directly, which further encourages this style of zero-shot prompting.

Prompting assumes access to a large language model (LM) pre-trained on generic text. An LM is a probabilistic model over text. Given an input text $\mathbf{x}$, it gives the probability of output text $\mathbf{y}$. The idea of the prompting technique has been facilitated by recent improvements in these models, primarily deriving from scaling Transformer neural networks~\cite{vaswani2017attention}. Recently, researchers have trained LMs that are directly targeted for the end-use of prompting~\cite{sanh2021multitask,wei2021flan}. These language models can all be queried in a standard way. In this work, we utilize three different LM queries (\autoref{fig:formulas_explained}): \vspace{1em}

\noindent
\textbf{Generation} - Sample a target output for a given input, 
\vspace{-.5em}
\[\tilde{\mathbf{y}} \sim p(\mathbf{y} \ |\ \mathbf{x}). \]

\noindent
\textbf{Ranking} - Compare the rank score of different texts, where $f$ is a function based on $p(\mathbf{y}\ | \ \mathbf{x})$(~\cite{brown2020gpt3}, details in \autoref{sec:implementation}), 
\vspace{-.5em}
\[f(\mathbf{y}^1, \mathbf{x}) <  f(\mathbf{y}^2, \mathbf{x}). \]

\noindent
\textbf{Top-K} - Find the $k$ highest probability outputs from the model,
\vspace{-.5em}
\[\mathrm{topk}_{\mathbf{y}'} p(\mathbf{y}=\mathbf{y}'\ |\ \mathbf{x}). \]

We represent an NLP \textit{task} as a table of examples, each associated with a fixed set of fields and a label. As a running example, we consider the task of document topic classification. For this task, there is one field, the article text, and one label, the topic of the article. The document text might consist of, 

\begin{quote}
Authorities have halted oil export flows from the main pipeline in southern Iraq after intelligence showed a rebel militia could strike infrastructure, an oil official said on Saturday...
\end{quote}

\noindent whereas the corresponding topic label would be \textit{World}. 

Contrast the standard ML approach for this task to prompting. A standard approach would collect supervised labels for the task and train a model on these examples. In zero-shot prompting, we do not have access to a training set. Prompting facilitates ad-hoc models by converting each test example directly to a form natural language input to which a large LM can respond. A user introduces a prompt template and a set of answer choices. The prompt template describes how to map the example fields to an input string $\mathbf{x}$, whereas the answer choices describe how to convert potentials outputs $\mathbf{y}$ back to labels for the task.

We follow the work of PromptSource~\cite{bach2022promptsource}, where researchers introduced a format for describing  \textit{prompt templates}. 

\begin{quote}
In which section of a newspaper would the text appear? \{document\} 
\end{quote}

\noindent with \textit{answer choices} given as a dictionary of labels: 

\begin{quote}
[World, Sports, Business, Science and Technology].
\end{quote}

\noindent Utilizing the prompt template, we can construct an example prompt $\mathbf{x}$ for model conditioning and answer choices $\mathbf{y}^1, \ldots, \mathbf{y}^n$. Each of the choices can then be ranked under the model to provide an \textit{evaluation score} for the dataset. 

We also support other task formats such as multiple-choice tasks that allow different answers for each example depending on specific fields in the data set. 

\begin{quote}
\{question\}  Choose between the following answers: \\(A) \{ answerA \} (B) \{ answerB \} ... 
\end{quote}

\noindent Here the answer choices could either be the corresponding letters or the answers themselves. We discuss this decision in Section~\ref{sec:rc}.

Throughout, we assume a small set of labeled validation data for quantitative evaluation, which differs from the research on true zero-shot learning~\cite{perez:neuraips2021}. We note though that this is not enough data to attempt to automatically generate prompts directly. 
The main elements of prompting can be summarized as:
\vspace{1em}

\noindent \textbf{M1 - Prompt Template.} A user writes a prompt template consisting of a task description in natural language that utilizes the fields from the task in a situated context. This leads to the construction of the input $\mathbf{x}$ that is used for conditioning of the large LM.

\noindent \textbf{M2 - Answer Choices.} A user provides a dictionary of answer choices paired with the original labels for a given task that offer different possible output wordings $\mathbf{y}$  to be considered by the model. The underlying model uses ranking to determine which of these answer choices to select. The original label paired with this answer choice is then the classification choice selected.

\noindent \textbf{M3 - Evaluation.} A user can evaluate the current version of the system under a known prompt for a set of validation data. This step will provide a proxy score for how well the given wording of the prompt is at capturing the underlying task.

\section{Goals and Tasks}
\label{sec:goals}
We held discussions with our NLP researchers on the team to determine the functionality for a  minimal viable prototype. We imagined how our tool could enable our example personas journalist and analyst to work with prompting. In the following, we summarize the insights of these discussions. 

While prompting is a promising approach, it is still too work-intensive for many use cases. The problem is that performance is highly dependent on the specific wording choices for templates (\textbf{M1, M2}), which is reflected in a high variance in accuracy (\textbf{M3}). For example, previous work has shown that different choices of prompts often lead to a more than 10-point spread in task accuracy between the best and worse choices at stage \textbf{M1}, even though both were approved by human editors~\cite{sanh2021multitask}.

Brute force approaches for prompt search require a user to write a large set of possible prompts and validate them empirically on a task. However, these approaches are both computationally expensive and slow. An interactive tool should provide an approach for using fewer resources and allow fast iterations for prompt engineering.

\subsection{Goals}

The high-level aim of \model is to provide a better approach for prompt development by domain experts in terms of four targeted goals:

\noindent \textbf{G1 - Support a broad set of ad-hoc NLP tasks.}  It is essential that the system is generic, as the user may not know beforehand the nature of the end-user task. The tool should present a single interface for multiple different potential downstream tasks. In this way, a tool is not targeted specifically to model trainers but also to the end-user's application goals. 

\noindent \textbf{G2 - Faster and more informed prompt writing through feedback from data.} The process should let users target new language tasks that arrive during their projects. A tool should enable the user to develop prompts efficiently. It should also provide feedback on what effect prompt variations have. The goal is to make prompt search less automated by giving the user the human-in-the-loop ability to edit and construct prompts based on their domain knowledge.

\noindent  \textbf{G3 - Ground prompt choices in quantitative measures.} Prompt customization replaces training for ad-hoc systems, but it is still critical that choices be grounded in task evaluation metrics. A key element of interaction in the system is that task scores be directly available in the tool itself.

\noindent  \textbf{G4 - Ease deployment of models to end uses. } The tool should provide a testbed for the wording and usage of the prompts, but for actual usage,
the prompts must be able to be run and used on actual data cases. Our goal is for the 
user to be able to directly export the prompts written in the system to a full system for use on the 
final task itself.

\subsection{Tasks}

Given these goals, we identified a series of tasks that guided the development of \model :

\noindent  \textbf{Task 1 - Formulating and trying out prompts and prompt variations.} 
To allow users to quickly explore and run different prompt templates for a specific NLP task, the tool provides an interactive interface for formulating and trying out many different prompts in a manner that provides feedback on a small set of real data examples. This interface is agnostic to the NLP task. [G1, G2] 

\noindent  \textbf{Task 2 - Encoding prediction details of the model.} 
Even after finding a good prompt template, it is critical to ensure that the answer choices provided ensure that the prompt is useful and leads to successful results on realistic NLP problems. The tool exposes the predictions of the model beyond ranking to allow for other choices that could be selected.
[G2, G3]
    
\noindent  \textbf{Task 3 - Testing promising prompts on task performance.} 
From Task 1 and Task 2, there are many combinations of prompt templates and answer choices to be considered, and each may be run on many different examples. Testing provides insight to the user on how choices for promising candidates lead to different resulting outcomes in a larger data regime and informs them which elements of their design have been successful.  [G3]

\noindent  \textbf{Task 4 - Export prompt for concrete deployment.} 
The end goal of the system is to provide packaged prompts that can be used in real tasks after the exploration phase has concluded. For this to be useful, the final system must collapse the exploration steps and provide a full prompt for deployment.  [G4]

\begin{figure}[tb]
    \centering
    \includegraphics[width=.9\linewidth]{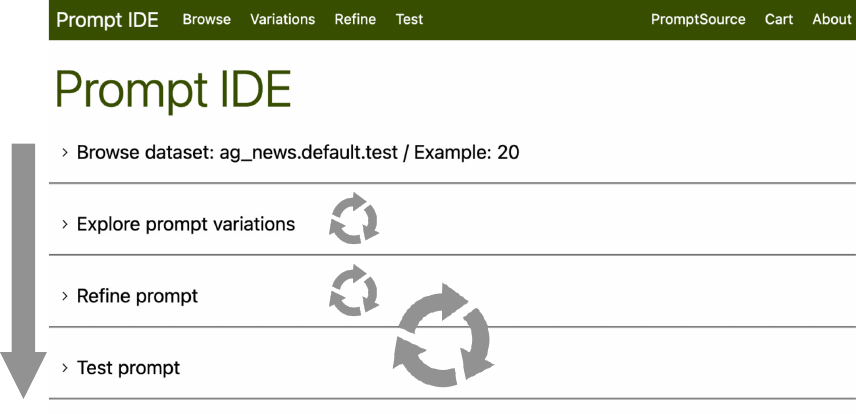}
    \caption{\model UI is organized like a notebook with foldable sections that follow the order of the main workflow but also allow quick iterations within a section or between neighboring sections. }
    \label{fig:ui_all_folded}
\end{figure}

\begin{figure}[tb]
    \centering
    \includegraphics[width=\linewidth]{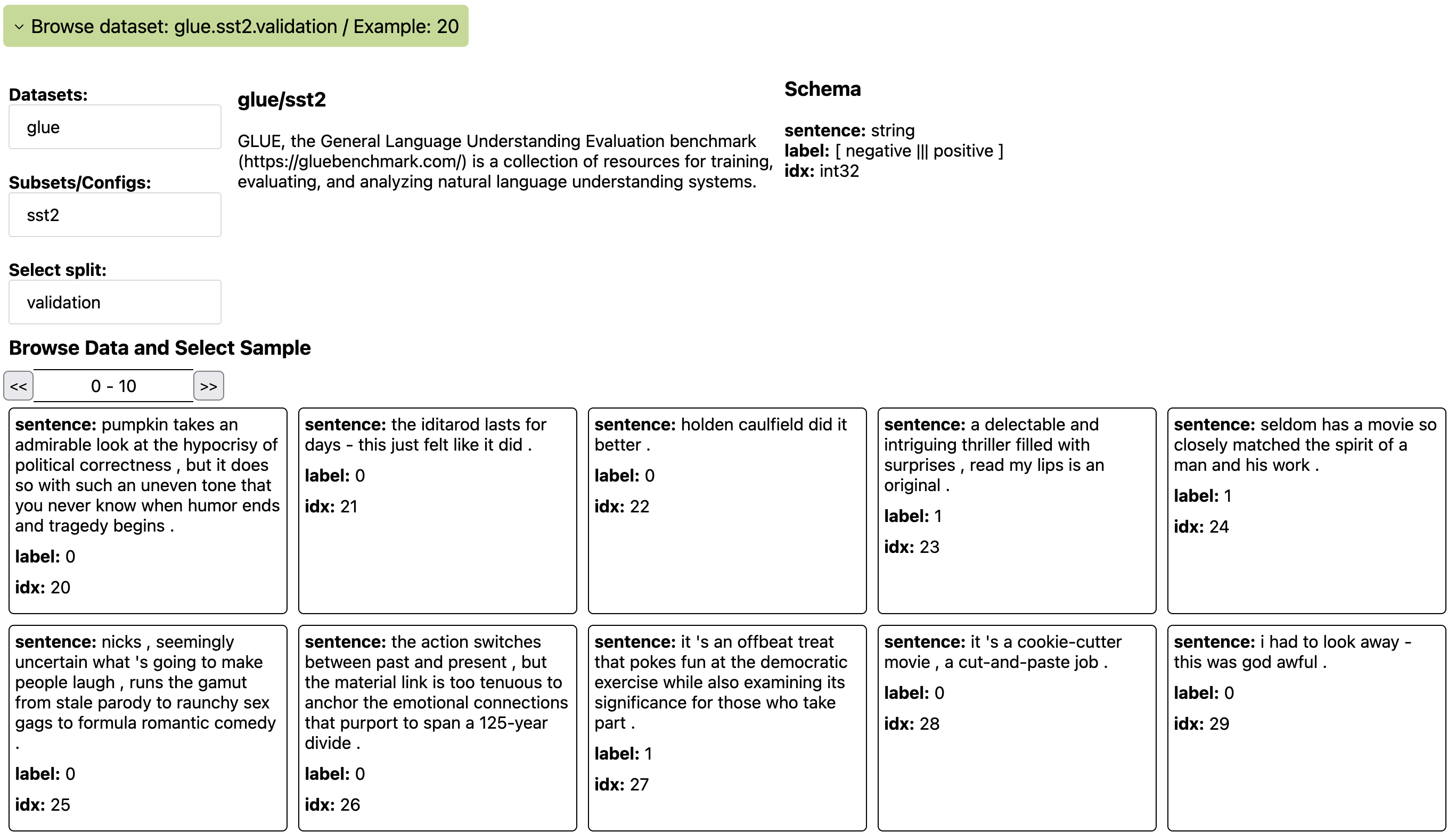}
    \caption{Dataset navigation. Browsing through dataset samples and dataset schema lets the user get acquainted with the data. In this case, ``label'' refers to the index position within a list of potential labels.}
    \label{fig:ui_dataset_browsing}
\end{figure}

\section{Design} 
\label{sec:design}

The visual interface of \model provides means to address the aforementioned tasks. With the text being the main carrier of information for the tasks, we designate the
most screen real estate to text and interaction with text while introducing visual encodings for abstraction when useful. 

At a high level, \model appears as a continuous notebook of four foldable sections (\autoref{fig:ui_all_folded})
that lead to the following workflow: a \textit{dataset navigation section} to select and browse data, a \textit{prompt variation section} to broadly explore prompt variations, a \textit{prompt refinement section} to help fine-tune a specific prompt, and a \textit{prompt testing section} to explore results of testing on a larger scale.

\subsection{The Four Sections of \model}

The \textbf{dataset navigation section} (\autoref{fig:ui_dataset_browsing}) enables browsing and selection of a reference dataset needed for testing prompt templates against tasks T1--T3 and respective goals G1--G3. It provides access to many standard NLP data using the huggingface datasets\cite{lhoest-etal-2021-datasets}. In addition, a user can provide their own data as CSV or JSON files. While describing the dataset schema is syntactically sufficient, the tool enables browsing to help get a better understanding of the encoded semantics behind each data dimension. E.g., in \autoref{fig:ui_dataset_browsing}, the dataset \textit{glue/sst2} (Stanford Sentiment Treebank v2) is selected, and the schema indicates the presence of three fields: \textit{sentence}, \textit{label}, and \textit{idx}. The naming of fields and the respective datatypes match the intuition of how a sentiment classification dataset could look like. But browsing shows that sentences can be very short and of low quality. 

\begin{figure}[tb]
    \centering
    \includegraphics[width=\linewidth]{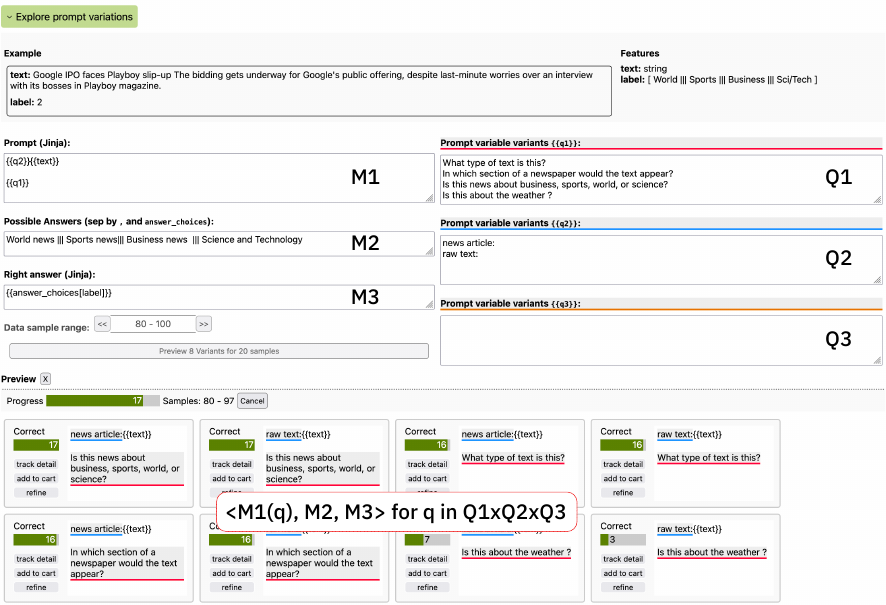}
    \caption{Prompt variation section. The user can quickly generate and progressively test variations of prompts to identify promising candidates. Each prompt variation is represented as a template card highlighting the values for template variables $q_x$ in unique colors and showing the number of correct answers vs. data tested.}
    \label{fig:ui_prompt_variations}
\end{figure}

\begin{figure*}[tb]
    \centering
    \includegraphics[width=.7\linewidth]{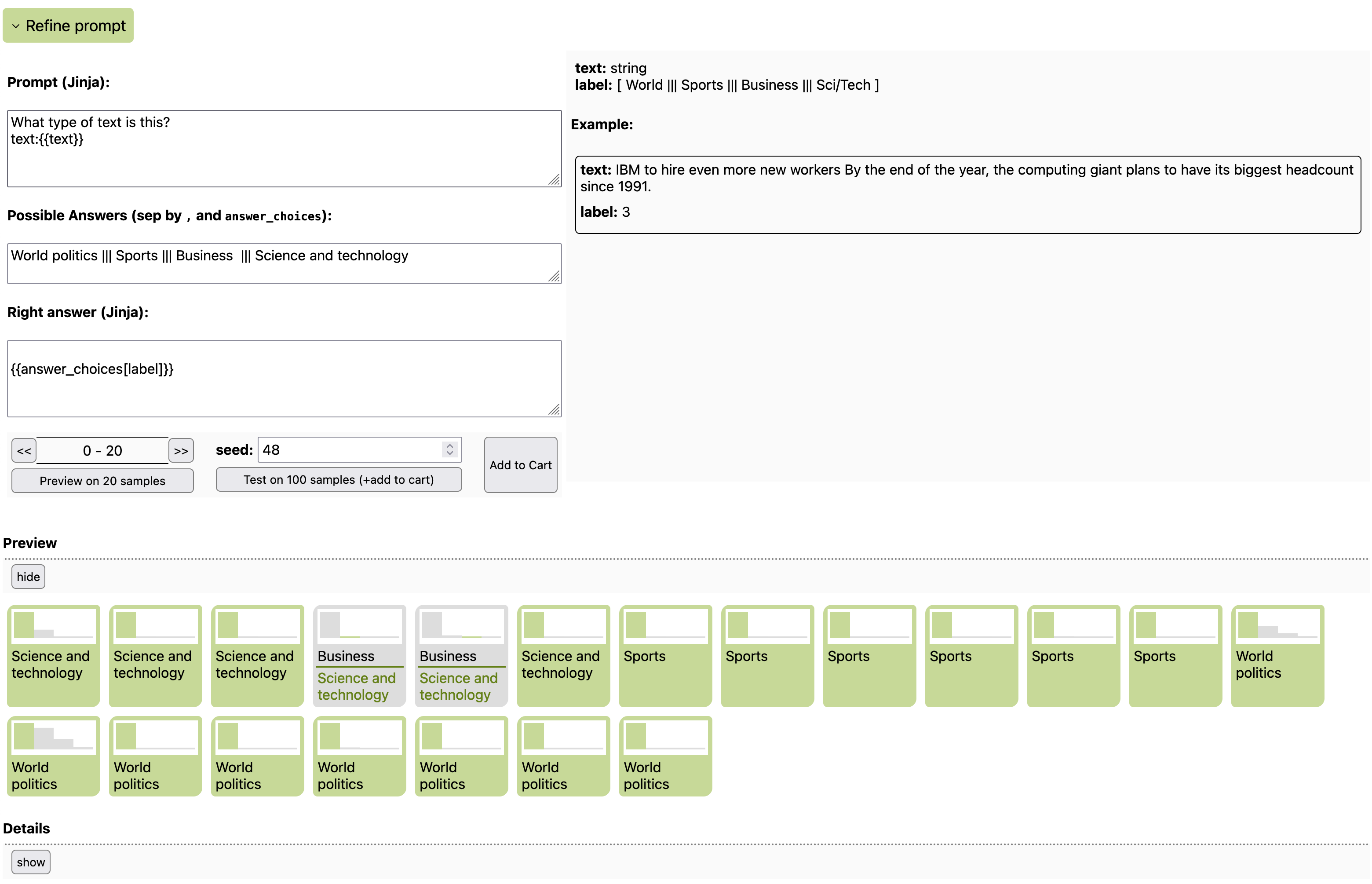}
    \caption{Prompt refinement section. To incorporate small optimizations for promising prompts, the user can test against a small subset of data frequently. For each data item, the evaluation chip indicates the prediction, the ground truth, if they match (green or gray), and the normalized distribution of probabilities as a bar chart on top. }
    \label{fig:ui_refinement}
\end{figure*}

The \textbf{prompt variation section} (\autoref{fig:ui_prompt_variations} and \autoref{fig:teaser}) allows formulating a prompt experiment for broad variations of prompt templates (M1 in \autoref{sec:model}) using up to three template variables $q_1, q_2, q_3$, and spanning their combinatorial space $Q1 \times Q2 \times Q3$. The user can formulate a prompt template (M1) using dataset fields (e.g., \textit{\{\{text}\}\}), template variables $q_x$, and plain template text. 
The list of answer choices (M2) can be formulated as a static or dynamic list (e.g., \textit{World $\vert$ $\vert$ $\vert$ Sports $\vert$ $\vert$ $\vert$ Business}). 
The correct answer (M3) can be dynamically retrieved from the dataset and the answer choices (e.g., \textit{answer\_choices[label]}). 

After defining the space of all variations $<M1(q), M2, M3>$ for $q \in Q1 \times Q2 \times Q3$, the model can now be asked to predict answers for all variations on a small set of data items. As soon as the experiment is started, each variation of $M1(q)$ is represented as a template card, highlighting the template variables $q_x$ by a preassigned color ($q_1$ in red, $q_2$ in blue, $q_3$ in gold) and showing the sum of correctly evaluated samples as a bar chart. Then, progressively, the full set of variations is tested against the set of data items, adding results for one data item at a time. For each step in the progression, the order of template cards is updated, keeping the stack sorted by decreasing performance against the ad-hoc task. At any time, when the user has gathered sufficient evidence of what could be promising candidates, they can stop the progression and re-iterate before the experiment would have been finished. This procedure of iteratively formulating prompt variations and trying them out on a small dataset addresses task 1 (G1, G2).

The \textbf{prompt refinement section} (\autoref{fig:ui_refinement}) takes one of the template variations and enables quick iterations for fine-tuning  this template. Using only one variant $M1$ allows the user to try on one batch of data items at interactive rates (T1, G1, G2). For each iteration, a performance overview is shown in an evaluation chip that indicates if the task was evaluated successfully for a data item (green background) or not (grey). It shows the predicted answer in black and the correct answer in green. If they match, only one is shown. On top of each evaluation chip, a bar chart indicates the relative probability of all possible answers sorted by rank and normalized to maximum probability.

The green bar highlights the ground truth, and the leftmost bar indicates the current prediction. This encoding allows, e.g., insights if a wrongful prediction was close to being a coin flip (similar height for most left bar and green bar) or if a correct prediction was a good choice (significant difference between the leftmost bar in green and the second bar from the left), targeting task T2. Besides running quick iterations using a small data regime, the user can trigger an experiment of testing against a larger test set (T3).

\begin{figure*}[tbh]
    \centering
    \includegraphics[width=.9\linewidth]{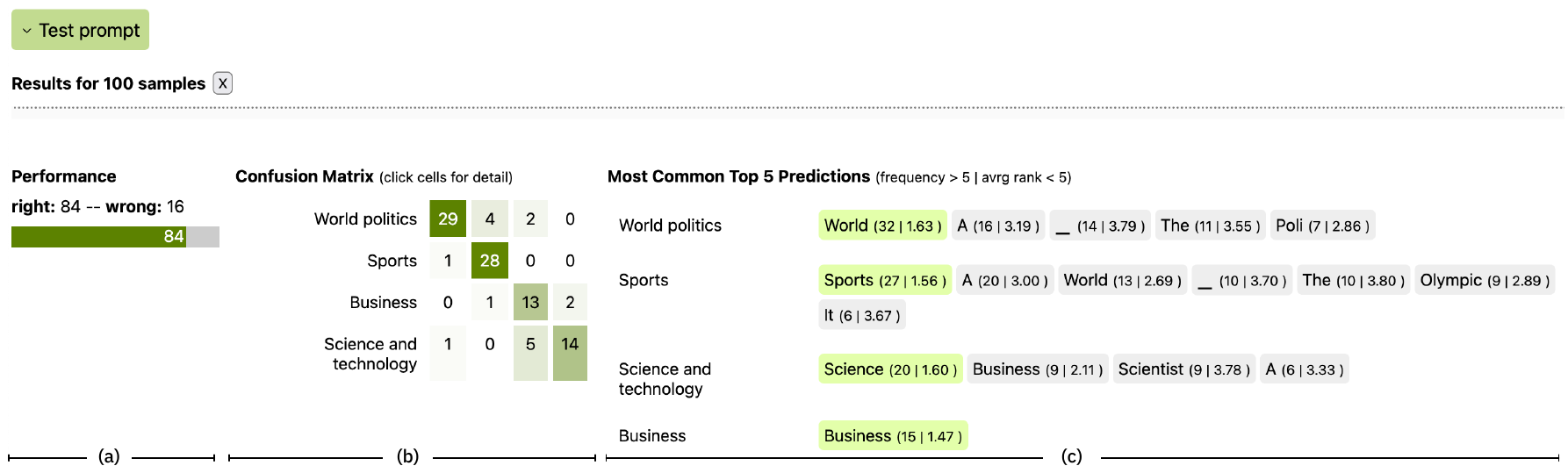}
    \caption{Prompt testing section. Showing results of testing against a mid-sized subset of data to get quantitative measures to help answer questions: How well did my prompt perform against the task (a)? What did the model confuse using my prompt (b)? How could I potentially tweak the answer choices (c)?}
    \label{fig:ui_testing_01}
\end{figure*}

The results of this testing are shown in the \textbf{prompt testing section} (\autoref{fig:ui_testing_01}) of \model. After the model completes the testing against a larger dataset, the results are presented such that a user might be able to answer the questions: How well did my prompt perform against the task (T3, G3)? What did the model confuse using my prompt (T2, T3, G3)? How could I potentially tweak the answer choices (T3, G3)?

For all three questions, \model provides a visual encoding that can help find an answer. A stacked bar chart indicates the share of correct predictions vs. incorrect ones to indicate prompt performance at the highest abstraction level (\autoref{fig:ui_testing_01}(a)). If the answer choices allow, a confusion matrix shows across class scores ((\autoref{fig:ui_testing_01}(b)) and if the in-class performance was good (large values on the diagonal axis). If the answer choices are dynamic but still form groups, the tool shows the top ten most abundant ground truth groups in the confusion matrix. For datasets where the answer choices do not fall into groups, nothing is shown.

Finally, to help answer the question about potential answer choice tweaks, the tool records the top five ranked generation tokens for each data item independently. Then, for each group of answer choices, these tokens are accumulated, and the number of appearances in the top five is recorded. Additionally, the average rank they had per answer group is calculated (value between 1 and 5). The list of the most frequent and highest-ranked answer tokens is shown in \autoref{fig:ui_testing_01}(c), with tokens sorted by decreasing appearance count. The green background indicates the best average rank in the group (which does not need to be the most appearing item). The use case in \autoref{sec:prompt_improvements} gives an example of the practical use of this feature. Similar to the confusion matrix, nothing is shown if the answer choices do not form groups. 

\subsection{Visual Encoding and Interactions}

Across the foldable sections, a user can investigate more detail about the prompted data items using the detail stripes (\autoref{fig:teaser}). Each detail stripe consists of an upper part highlighting the answer options, the predicted answer (on top), and the ground truth answer (encoded in green). The panels below show details about the prompted text, the ground truth answer, and what the model has generated. Upon request, each ranked result can also show the probability of the respective answer option.

Detail stripes can be shown when tracking a specific template variation (\autoref{fig:ui_prompt_variations}) in the prompt variation section. They can be unfolded beneath the performance chips in the prompt refinement section (\autoref{fig:ui_refinement}). They show detail about the respective subsets when the user clicks on the performance bars or the cells of the confusion matrix in the prompt testing section (\autoref{fig:ui_testing_01}). Detail stripes target task T2 (G2).

\begin{figure}[tbh]
    \centering
    \includegraphics[width=\linewidth]{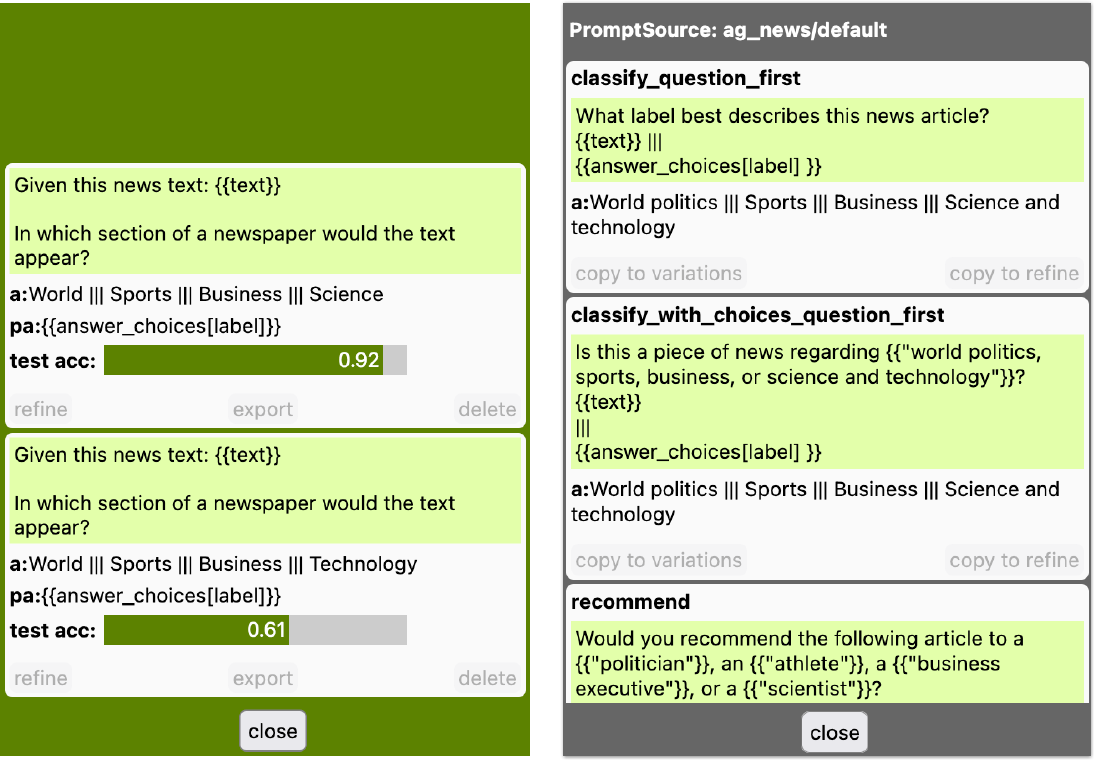}
    \caption{Shopping carts allow users to collect their own prompt variants (left) or browse through prompts made available by the BigScience promptsource community (right).}
    \label{fig:ui_shopping_carts}
\end{figure}

To collect and store promising prompts, \model provides a shopping cart (\autoref{fig:ui_shopping_carts}). From the cart, templates can be sent to the prompt variation section or the prompt refinement section. Templates can be exported from the cart to deploy them seamlessly with the tooling provided in the \model repository (T4, G4). To add prompts, a user can trigger buttons from the template cards in the prompt variation section (\autoref{fig:ui_prompt_variations}) or the refinement section (\autoref{fig:ui_refinement}). If a prompt has been evaluated against the larger testing set, the performance result will be automatically added to the corresponding item in the cart for comparison. Furthermore, a read-only PromptSource shopping cart (\autoref{fig:ui_refinement}) shows templates that have been created for a specific dataset by the global community. These prompts could serve as inspiration for starting a user's own prompt idea.

\subsection{Example Interactive Workflow}

A prototypical interaction workflow using \model starts with opening the data browser section to investigate the schema and concrete examples of own or globally available data. The user clicks on one item that serves as a good exemplar and is shown for reference in the prompt variations and prompt refinement sections. The user then opens the prompt variation section and writes down a prompt template using data and template variables. Alternatively, they open the prompt source cart and scan for examples that could serve as inspiration, copy them over, and add template variables. They run the experiment and stop it early because one prompt variation seems to perform very well. They use the shortcut to copy the specific prompt over to the refinement section, where, through multiple small edits and along multiple data portions, the prompt seems to be performing well. During this, the users observe the performance chips and occasionally the detail stripes. Later, the user triggers the larger-scale testing to see if they over-optimized the prompt for a local data range. After a short period of time, they note a high confusion between labels A and D. Inspired by the most common top five predictions, they iterate over their answer choices and run the test again, resulting in better performance. The prompt (and maybe some intermediate steps) are saved in the shopping cart. The best-performing prompt is exported to a JSON file. The user can now run the newly created ad-hoc model (original LM plus prompts in the JSON file) with a simple input-output interface or as a batch processing script on new data for their customers. 

\begin{figure}[tbh]
    \centering
    \includegraphics[width=\linewidth]{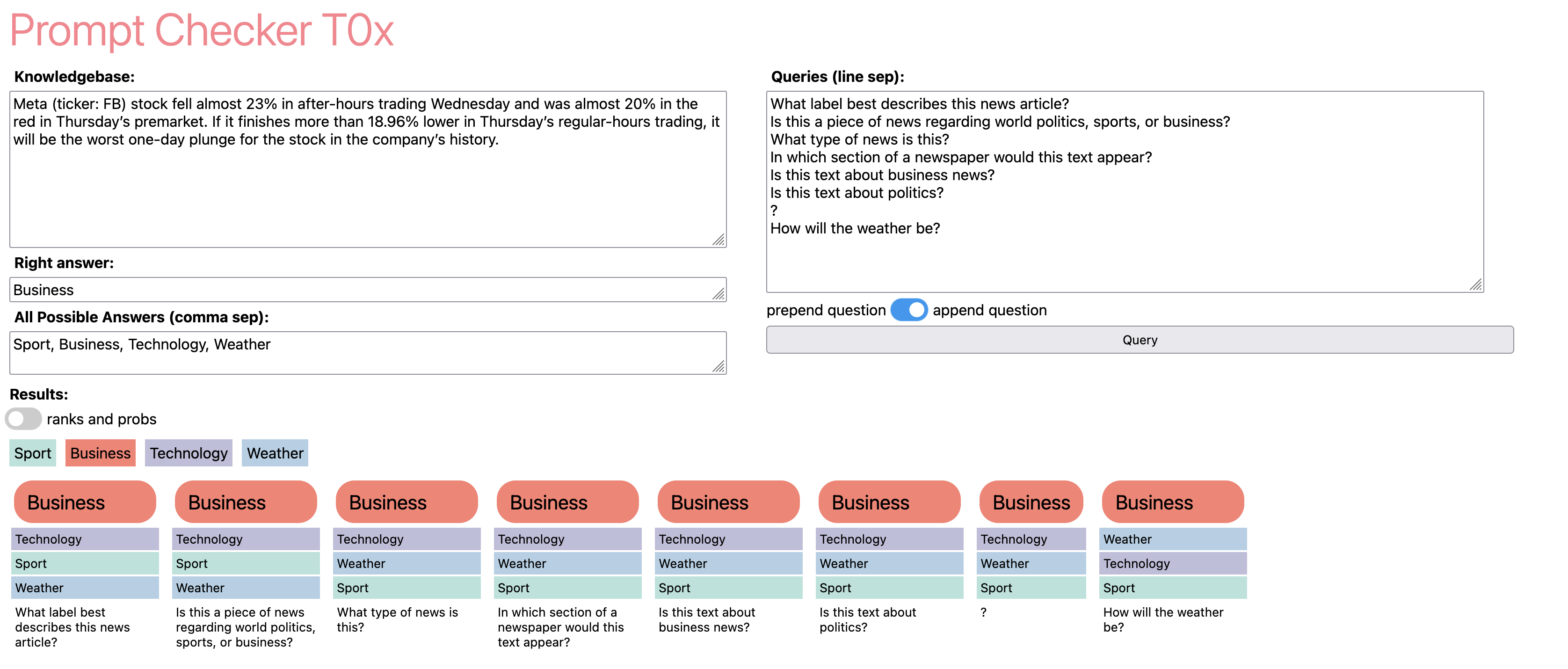}
    \caption{Early version of prompt variation testing. The interface is less powerful than the final version: only one template variable $q_1$ that can either be prefixed or appended to the prompt. Testing is only against one dataset item. Each answer option has its own color, which was not considered useful by early users. }
    \label{fig:ui_early_iteration}
\end{figure}

\subsection{Design Iterations and Rationale}
\label{sec:design_iter}
While consulting with our co-authors, who are NLP domain experts, we went through multiple design iterations on different parts of the tool. In this section, we highlight some of them and provide a more in-depth design rationale for certain parts of \model. 

The overall design as a notebook with foldable sections was inspired by the popular use of Jupyter notebooks in the NLP community. It allows occupying the screen with different views while keeping the views connected in a natural order. This allows the user to build a mental map that is established by scrolling up and down. If, instead, we had mapped the steps of the interactive workflow to independent views, the user might not be able to build this mental model due to the constant context switches. To assist with navigating to a specific subsection, the menu bar acts as a table of content that scroll-animates to the respective sections. 

\begin{figure*}[ht]
    \centering
    \includegraphics[width=\linewidth]{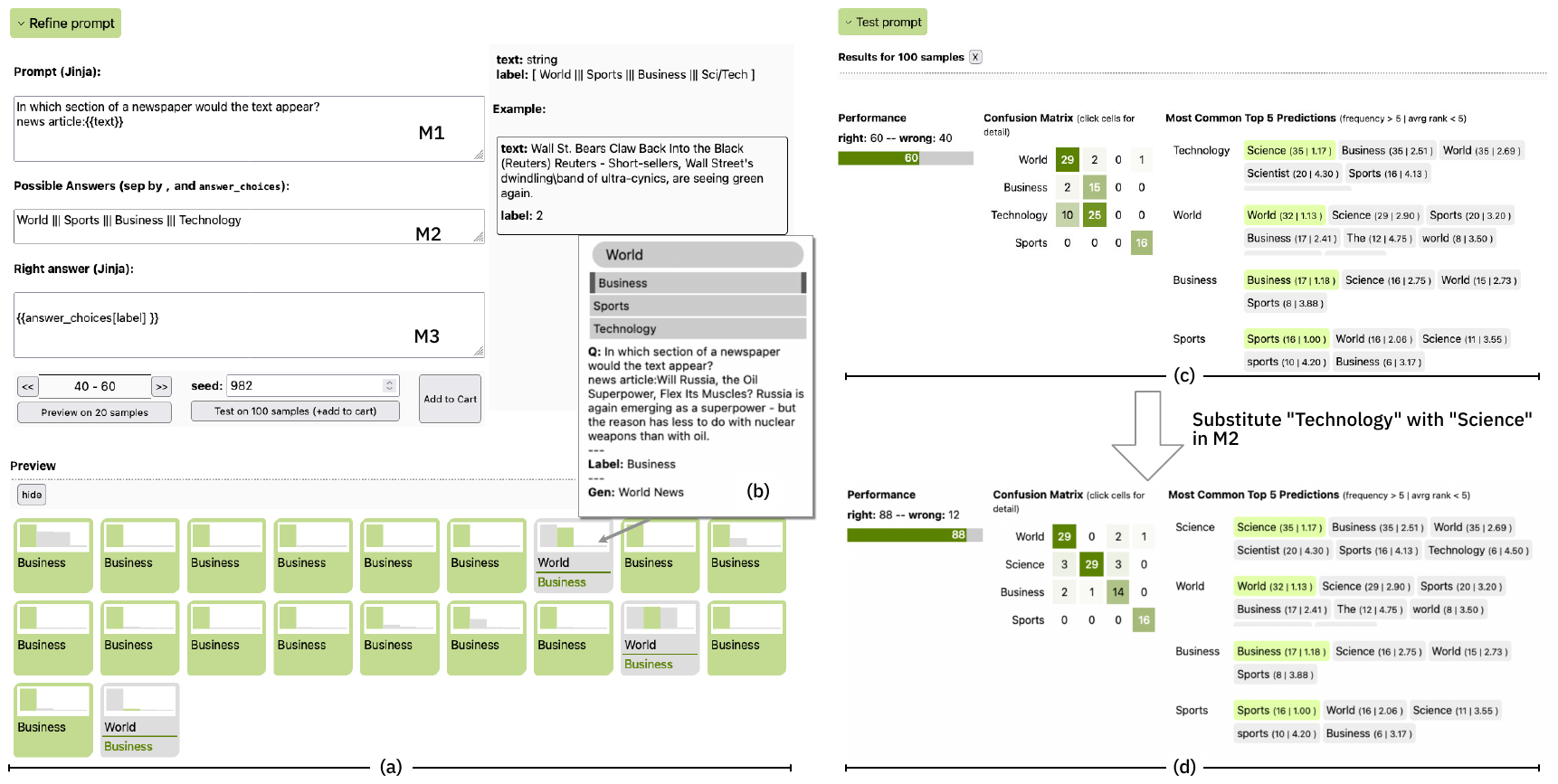}
    \caption{Use Case Document Classification for a news dataset. Details in \autoref{sec:prompt_improvements} }
    \label{fig:usecase1}
\end{figure*}

Our use of progressive visualization methods~\cite{progressive_arxive} was driven by technical design considerations. We started thinking about what the modes of interaction during a progression would be. We decided to use progressive updates for the prompt variation testing because early stopping in this phase of the exploration has proven useful. For example, testing on data with just one label class may reveal local effects relatively quickly, in which case the user can stop the test. On the other hand, we did not use the early stopping interaction during prompt testing on a larger dataset. The goal for this part of the workflow is to observe more global effects by running the test to completion. Early stopping might increase the chance of observing a subset of the data that has local effects, negating the intention of larger-scale testing. Furthermore, the progression in the prompt variation section is controlled by the client and stops if the tool is closed. The larger-scale testing is controlled by the server and continues running even when \model is being closed on the client-side.

A fundamental technical design decision was which language to use for formulating the prompts with data and template variables (for T1). The three options for our decision were: (1) invent a new language, (2) use a templating language, or (3) use a general programming language. We quickly decided against inventing our own language that would have to be explained and maintained. The choice for using a templating language (Jinja) over a general programming language (Python) was based on the observation that PromptSource~\cite{sanh2021multitask} used the same language. This made the PromptSource parsing work immediately available to us and also allowed us to build the read-only shopping cart with prompt examples gathered by the partner project. Another concern was that a general programming language could be a prime target for malicious server attacks once the tool is released to the public.

The detail stripes underwent several design iterations. An early version assigned a color to each answer choice (\autoref{fig:ui_early_iteration}). But our domain experts did not express interest in being able to track ranks of answer choices across data items and found it more distracting than useful. The switch to a simpler color encoding was also enforced by the required capability to allow dynamic answer options that change per data item and repeat a few times. This would have required an impractically large number of categorical colors.

\subsection{Implementation}
\label{sec:implementation}
Our \model prototype consists of a backend in Python that communicates with a frontend written in Typescript and Svelte. We use the openly available pre-trained T0-3B large language model~\cite{sanh2021multitask} provided through the huggingface platform~\footnote{\url{https://huggingface.co/bigscience/T0_3B}}. For long-running queries, the backend provides a custom-built queuing and execution system that keeps the memory footprint for the model low. In our implementation, answer options are ranked by the decreasing average log-likelihood: 
\vspace{-.5em}
\[ (\sum^{i<l_a}_{i=0} \log(p_a^i))/l_a, \]

\noindent
with $p_a^i$ being the probability for the $i$-th token of answer $a$ and $l_a$ being the answer's token length. 

The demo system is available at \url{http://prompt.vizhub.ai}. We will make the source code available upon acceptance of the paper. The open-source version allows easy use of custom data that is either provided as a CSV file or by the Huggingface dataset interface.

\section{Use Cases}
\label{sec:usecases}

We illustrate how we can use \model to interactively prompt a diversity of NLP tasks, compare these prompts, save them and export them outside of \model. We experiment with a range of standard tasks in NLP, including document classification, reading comprehension, and natural language inference. The tool enables seamless development of prompts for a wide variety of tasks and formats [G1] while quickly providing feedback on prompts patterns that generalize to many instances [G2]. 

\subsection{Document Classification}
\label{sec:prompt_improvements}

The most common end-user task for NLP~\cite{DBLP:conf/emnlp/LhoestMJTPPCDPT21} is document classification for an end-domain.  The task of document classification is to determine the label of a document from a fixed set. It can be used in ad-hoc models for filtering a large set of documents or collecting statistics about a large collection. Domain expertise is critical in classification since the specific wording may lead to different results. 

As an example use case we consider a prototypical version of this task with the goal of classifying the topic of a document. The AG News dataset~\cite{Zhang2015CharacterlevelCN} is often used for benchmarking this task and consists of text documents and labels that indicate which topic they are associated with (labels are canonically listed as ``World, Sports, Business, Sci/Tech''). This task (introduced in~\autoref{sec:model}) is representative of an ad-hoc classification task that a user might consider for prompt development with a language model [G1]. 

We can first explore this dataset through the dataset navigation section and then initialize the process through the prompt variation section. This section allows the user to write several different prompt templates [T1] as well as answer templates for these prompts [T2]. \autoref{fig:teaser} (a) shows some examples of these templates and the way they use fields from the underlying data set and template variables \texttt{q1} and \texttt{q2}. We select one of the better-performing prompts to investigate further in the prompt refinement section. 

Figure \ref{fig:usecase1}(a)  shows the output of the selected prompt for some data, and we observe that there is some confusion about the labels. Upon detailed inspection (see example in  \autoref{fig:usecase1}(b)), we observe that the labels can be ambiguous even under human evaluation. So, we send the prompt to larger-scale testing [T3,G3].

The testing reveals the specific problem. The ground truth label ``Technology'' is not a great choice and gets confused with other phrases (\autoref{fig:usecase1}(c)). Upon inspection of the Top 5 Rank Predictions, we observe that the token ``Science'' is a very frequent wording for this ground truth, even more than the term ``Technology'' itself. We use this insight to refine the answer options in the refinement section by substituting ``Technology'' with ``Science''. In this case, the wording of the answer changed, but other feedback would lead to changes in the wording of the prompt itself.

We run the testing again. After receiving the results, we see that the modification increased performance substantially (\autoref{fig:usecase1}(d)). We can now go to the shopping cart and export the new model [T4,G4].

\subsection{Multiple-Choice Answering}
\label{sec:rc}

The task of reading comprehension is to answer a question from a complex document. It can be used for ad-hoc models to find documents that provide answers to specific questions. The RACE dataset~\cite{lai-etal-2017-race}
is a multiple-choice reading comprehension dataset built from English examinations used for benchmarking this task. For a given text extract and a question about that text, the model has to choose the correct answer among four possibilities. Unlike document classification, the possibilities are different for each sample.

\begin{figure}[tb]
    \centering
    \includegraphics[width=\linewidth]{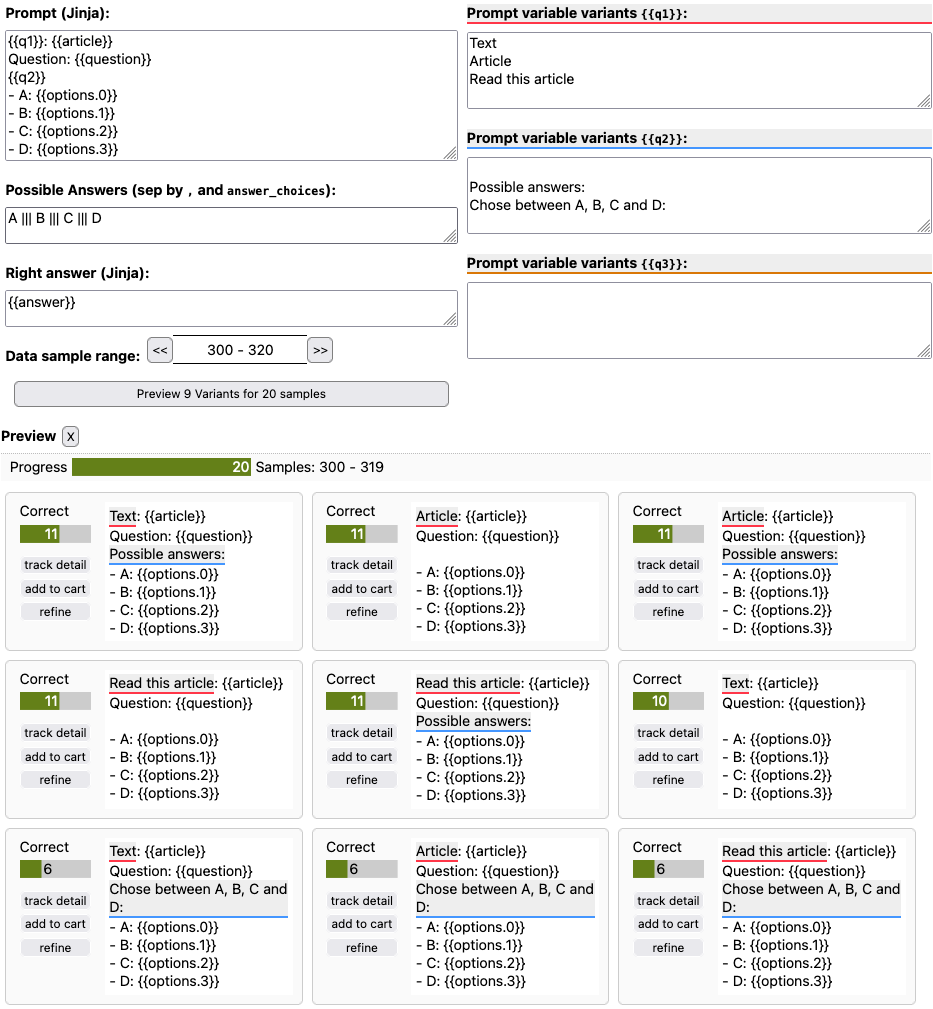}
    \caption{Mapping dynamic answer choices to simple outputs such as letters can help the model by simplifying its output space.}
    \label{fig:reading_comprehension_choose_bad}
\end{figure}

We explore prompts that introduce four answer choices, associating them with letters (A, B, C, and D): \textit{``Possible answers:''}, \textit{``Choose between A, B, C and D:''}, or nothing. Figure~\ref{fig:reading_comprehension_choose_bad} shows that \textit{``Choose between A, B, C, and D:''} consistently gives worse results than the two other variations, independently of how the input is introduced (\texttt{q1}). We discard this variation, and following~\cite{wei2021flan}, we try out prompts that explicitly state the instruction at the beginning of the prompt. We note that even though the performance remains the same for most of the prompt variations, some variations degrade the performance (for instance \textit{``Please select the correct answer among all of the options.''}).

\begin{figure}[tbh]
    \centering
    \includegraphics[width=1.0\linewidth]{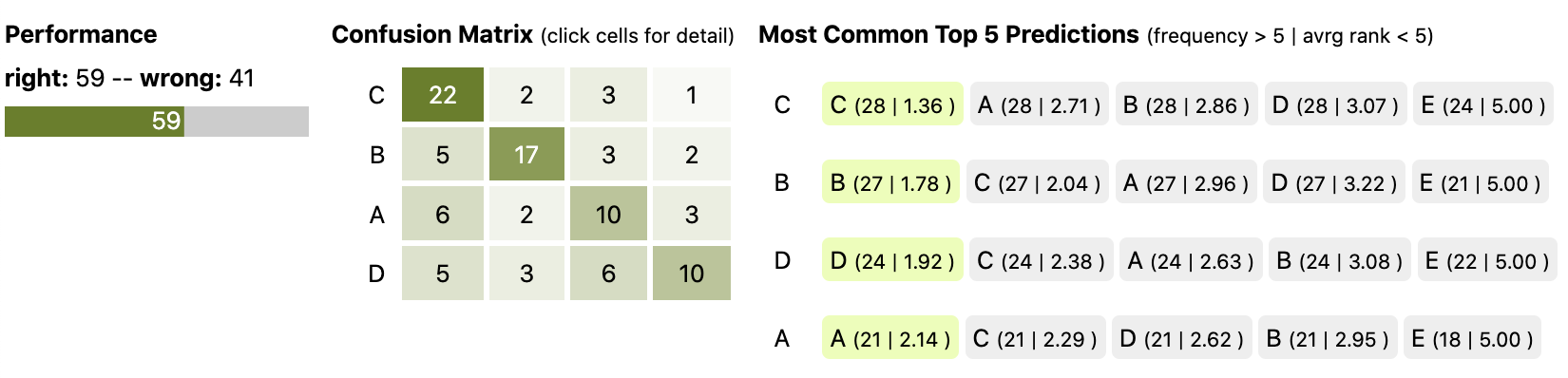}
    \caption{The label \textit{``C''} is systematically the second most frequent prediction of the model, which hints at a potential class imbalance in the training set of the underlying model.}
    \label{fig:confusion_matrix_class_C}
\end{figure}

We select one of these prompts for refining and test it on 100 examples. From the confusion matrix and the most common top 5 predictions (Figure~\ref{fig:confusion_matrix_class_C}), we notice that the model is often predicting \textit{``E''} even though it is not among the answer choices and that it tends to predict \textit{``C''}, as it is always the second most frequent prediction when it is not the correct label. This observation hints at potential class imbalances in the training mixture of the underlying model and warrants more investigation. We find that the training mixture contains a variety of multiple-choice question answering datasets that contain from 3 (COS-E~\cite{rajani2019explain}) to 8 (QASC~\cite{qasc2020}) answer choices, which leads to the high frequency of labels A, B, and C.

The interactive nature of \model makes it easy to develop a lot of prompts, save them and export them to a different environment. Once the best prompt has been identified, the end-user can use them outside of \model, for instance, to deploy a prompted language model in production [G4]. The JSON export format makes the whole interactive development extremely versatile and compatible with most of the standard codebase.

\subsection{Sentence Similarity}
\label{sec:NLI}

\begin{figure*}[tbh]
    \centering
    \includegraphics[width=\linewidth]{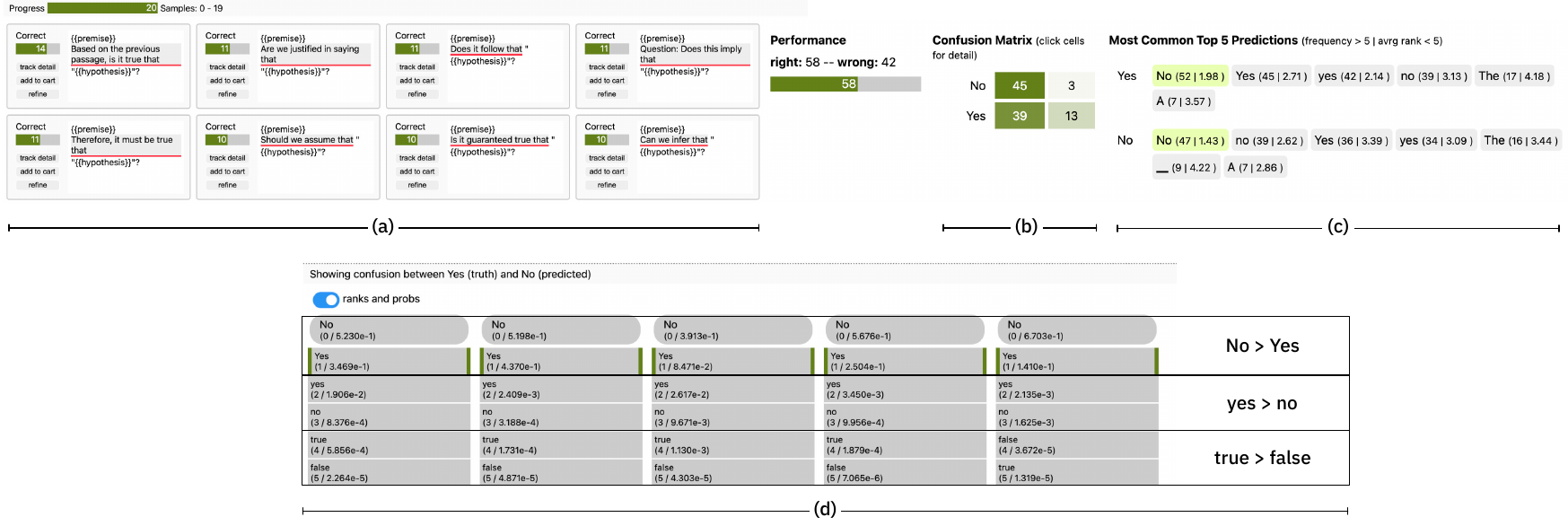}
    \caption{Use case Natural Language Inference on RTE dataset~\cite{RTE}. Details in \autoref{sec:NLI} }
    \label{fig:usecase_robuts_01}
\end{figure*}

\begin{figure*}[tbh]
    \centering
    \includegraphics[width=\linewidth]{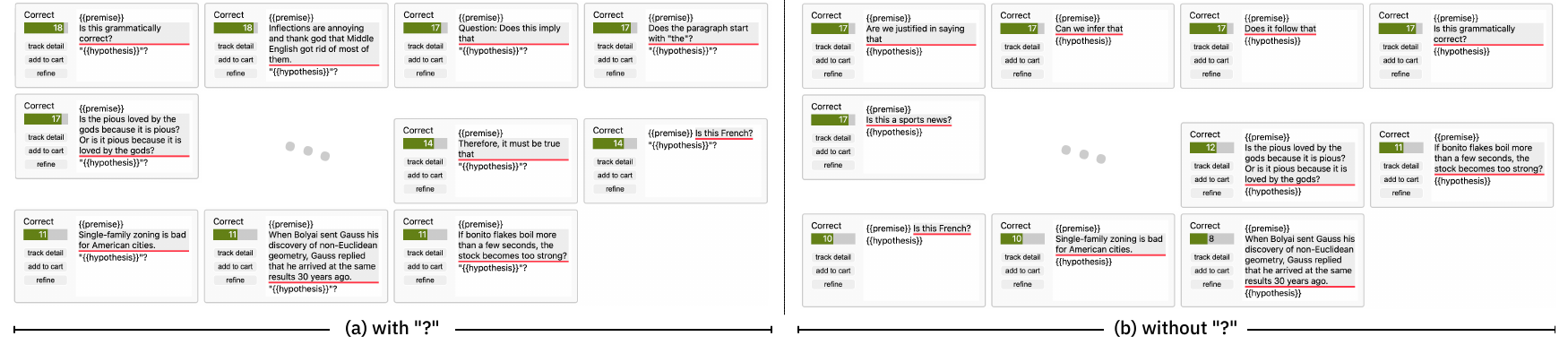}
    \caption{Using \model to study adversarial conditions by using misleading-extreme prompts from~\cite{albert:arxiv2021}. Details in \autoref{sec:NLI}}
    \label{fig:usecase_robuts_02}
\end{figure*}

The task of natural language inference is to determine the semantic relationship between two similar sentences. An ad-hoc domain expert would use a similar task to check whether documents agree with or contradict specific target statements. 
Recognizing Textual Entailment \cite{RTE} is an inference dataset where a model is asked to assume that one piece of text (the “premise”) is true and to classify whether another piece of text (the “hypothesis”) must also be true, i.e., whether the premise “entails” the hypothesis. For example, if the premise is \textit{Steve Jobs was attacked by Sculley and other Apple executives for not delivering enough hot new products and resigned from the company a few weeks later.} and the hypothesis states that \textit{Steve Jobs worked for Apple.}, then the data is labeled as ``true'' or ``entailment''.

Past work~\cite{albert:arxiv2021} has used prompting for this task but found that it was difficult to find a prompt wording that performed robustly on different examples.  Starting from the prompts in \cite{albert:arxiv2021} 
we can use \model for estimating the models' robustness to different wordings in the templates. In particular, the prompt variable variants $(q1, q2, q3)$ allow us to easily control and test wording variations.

To start finding good prompts for RTE, we copy and paste the main instructive variations of \cite{RTE} into \texttt{q1}. In previewing 20 examples, we see T0-3B's performance across prompts ranges from a respectable 70\% to 50\% (\autoref{fig:usecase_robuts_01}a). We add the best and the worst prompts to carts for further investigation under the “Refine Prompt” section. From the confusion matrix (\autoref{fig:usecase_robuts_01}b), see that most mistakes are models under-predicting entailment for sentence pairs where the premises do, in fact, entail their hypotheses (i.e., ground truth is ``Yes''). %

Another noteworthy finding from the “Most Common Top 5 Predictions” is that, for the entailment class, the model generates “No” with a higher rank more frequently than “Yes” (\autoref{fig:usecase_robuts_01}c), yet it also generates “yes” and “True” with higher rank more frequently than “no” and “False” (\autoref{fig:usecase_robuts_01}d). Similar to \autoref{sec:prompt_improvements}, we could simply change the desired target words for a performance boost. But we can also study the effect of explicitly providing the answer choices in the input sequence by appending “True or false?” to every template, which further improves performance for all templates. However, explicitly providing the answer choices in input sequences does not always control model behaviors in accord with human intuition. Prepending the answer choices at the beginning of input sequences does not confer any performance boost.

Finally, \model is also prime for studying adversarial conditions such as irrelevant and misleading prompts. For example, using the misleading-extreme prompts from \cite{albert:arxiv2021}, \autoref{fig:usecase_robuts_02}a shows that “is this grammatically correct” outperforms just instructive prompts such as “does this imply that”. However, if we remove the question marks in the global template, then the misleading and irrelevant prompts consistently underperform the instructive ones (\autoref{fig:usecase_robuts_02}b), suggesting that models may be using question marks as some kind of heuristic feature. 

\section{Early Feedback and Lessons Learned}
After internal deployment within the prompting expert group we received initial qualitative feedback from two colleagues. When being asked about most helpful PromptIDE features, they answered: ``visualizing the best verbalizer in the confusion matrix form and the \textit{Most Common Top 5 Predictions} format. Furthermore, it's really helpful to preview the prompt variations and their performances and use \textit{refine} to copy to the Prompt Refine section.'' 

Main criticism came from the limitation of prompt variant variables: ``In agnews dataset, when I add q3 and q4, the Prompt variable variants field doesn't have q4. \dots we could have some indicator of maximum prompts allowed.'' Interestingly, there where also recommendations about UI improvements: ``I am confused about the use of red underlines \dots I would prefer to use something other than red because red indicates something is wrong.'' or simplifications like ``I think using some `i' popup icon for storing some information (with more descriptions) is much better than outright displaying technical texts like \textit{(frequency $>$ 5 $|$ avrg rank $<$ 5)}.''

The lesson we learned from working on this project are manifold with some of them are already indicated in \autoref{sec:design_iter}. The surprise to us was how much constructive feedback we got about the user interface during the development and after deployment, often started with an apology about being a non expert in UI. We recommend to establishing early on that all UI feedback is very welcome and not a overstepping in competences.

A constant fight for compromising between flexibility vs simplicity became a major task. E.g., determining how many prompt variation variables we should support (q1, q2, q3) vs how many can we handle with combinatorical explosion of variants.

Another lesson learned was that the more complex charts tend to be ignored whilst the simple charts with complex algorithmic ideas got much faster appreciated and accepted. E.g., the most common ``Aha!'' moment was caused by top 5 tokens idea in the prompt testing section.

\section{Conclusions and Future Work}
\label{sec:conclusion}

We present \model, a system for domain experts to customize models for ad-hoc tasks without requiring training expertise. The system adopts the prompting framework that has recently been developed for NLP tasks while developing an interactive visualization environment for customizing prompts for new tasks. The approach extends beyond brute-force prompt trial-and-error to facilitate the exploration of prompt language and the development of new prompt templates and answer choices. The system is open-source~\url{ http://prompt.vizhub.ai/} and can work with any available language model backend.

As the methodology of prompting develops, there are many areas of extension for \model. Currently, the tool supports tasks with a known set of choices, but there are many NLP tasks where the correct label may be a free-form response. These could be incorporated through support for new metrics such as BLEU in the prompt refinement section. \model also assumes that the user is able to deduce how to update prompts based on task metrics. Ideally, the system could provide direct advice on how to change prompts or highlight text spans that would lead to better results through methods like gradient saliency. These methods are currently too computationally expensive to run on large promptable models, but will likely improve with more research.  

\section{Acknowledgements}

We would like to thank the anonymous reviewers for their constructive feedback and helpful comments. This work was partially funded through NSF grant IIS-1901030 and the MIT-IBM Watson AI Lab.

\bibliographystyle{abbrv-doi}

\bibliography{template,ref,anthology}
\end{document}